# ParseTalk about Sentence- and Text-Level Anaphora

**Michael Strube** and **Udo Hahn**
$\mathcal{CLIF}$ – Computational Linguistics Research Group
Freiburg University
D-79085 Freiburg, Germany
{strube, hahn}@coling.uni-freiburg.de

**Abstract**
We provide a unified account of sentence-level and text-level anaphora within the framework of a dependency-based grammar model. Criteria for anaphora resolution within sentence boundaries rephrase major concepts from GB's binding theory, while those for text-level anaphora incorporate an adapted version of a Grosz-Sidner-style focus model.

## 1 Introduction

This paper treats the resolution of anaphora within the framework of *ParseTalk*, a dependency-oriented grammar model that incorporates strict lexicalization, head-orientation (based on valency specifications), feature unification, and inheritance among lexicalized grammar specifications (Bröker et al., 1994; Hahn et al., 1994). The results we present rest upon two major assumptions:

1. As many forms of anaphors (e.g., nominal and pronominal anaphors) occur within sentence boundaries (so-called intra-sentential or sentence anaphora) and beyond (inter-sentential or text anaphora), adequate theories of anaphora should allow the formulation of grammatical regularities for both types using a common set of grammatical primitives.

2. Anaphora are only one, yet very prominent phenomenon that yields textual cohesion in discourse. Adequate grammars should therefore also be easily extensible to cover non-anaphoric text phenomena (e.g., coherence relations, rhetorical predicates), which provide for additional levels of text (macro) structure, with descriptions stated at the same level of theoretical sophistication as for anaphora.

First, we will briefly compare our approach with work done in the context of government-binding (GB) grammar and discourse representation theory (DRT). As we conceive it, binding theory as developed within the GB framework (Chomsky, 1981; Kuno, 1987) offers one of the most sophisticated approaches for treating anaphora at the sentence level of description. This has also been recognized by advocates of competing grammar formalisms, who have elaborated on GB's binding principles (cf., e.g., Pollard and Sag (1992) within the context of HPSG, whose treatment is nevertheless restricted to reflexive pronouns). Interestingly enough, when faced with some crucial linguistic phenomena, such as topicalization, GB must assume rather complex movement operations in order to cope with the data in a satisfactory manner. Things get even more complicated when languages with relatively free word order, such as German, are taken into account. Finally, considering the case of text anaphora, binding theory has nothing to offer at all.

Another strong alternative for considering anaphora constitutes the framework of DRT (Kamp and Reyle, 1993). Its development can be considered a landmark in the model-theoretic semantic analysis of various forms of quantified sentences, conditionals, and anaphorically linked multi-sentential discourse. At this level of description, DRT is clearly superior to GB. On the other hand, its lack of an equally thorough treatment of complex syntactic constructions makes it inferior to GB. These deficits are no wonder, since DRT is not committed to any particular syntactic theory, and thus cannot place strict enough syntactic constraints on the admissible constituent structures. Focusing on the text analysis potential of DRT, its complex machinery might work in a satisfactory way for several well-studied forms of anaphora, but it necessarily fails if various non-anaphoric text phenomena need to be interpreted. This is particularly true of conceptually-rooted and pragmatically driven inferences necessary to build up textual macro structures in terms of coherence relations (Hobbs, 1982) or rhetorical structures (Mann and Thompson, 1988). This shortcoming is simply due to the fact that DRT is basically a semantic theory, not a comprehensive model for text understanding; it lacks any systematic connec-tion to comprehensive reasoning systems covering the conceptual knowledge and specific problem-solving models underlying the chosen domain.

Summing up, DRT is fairly restricted both with respect to the incorporation of powerful syntactic constraints at the sentence level and its extension to the level of (non-anaphoric) text macro structures. GB, on the other hand, is strong with respect to the specification of binding conditions at the sentence level, but offers no opportunity at all to extend its analytic scope beyond that sentential level. We claim, however, that the dependency-based grammar model underlying *ParseTalk*

1. covers intra-sentential anaphora at the same level of descriptive adequacy as current GB, although it provides less complex representation structures than GB analyses; these structures are nevertheless expressive enough to capture the relevant distinctions,
2. does not exhibit an increasing level of structural complexity when faced with crucial linguistic phenomena which cause considerable problems for current GB theory,
3. goes beyond GB in that it allows the treatment of anaphora at the text level of description within the same grammar formalism as is used for sentence level anaphora, and,
4. goes beyond the anaphora-centered treatment of text structure characteristic of the DRT approach in that it already accounts for the resolution of text-level ellipsis (sometimes also referred to as functional anaphora, cf. Hahn and Strube (1995)) and the interpretation of text macro structures (a preliminary study is presented in Hahn (1992)).

## 2  DG Constraints on Anaphora

In this section, we present, quite informally, some constraints on intra-sentential anaphora in terms of dependency grammar (DG). We will reconsider these constraints in Section 3, where our grammar model is dealt with in more depth. We provide here a definition of *d-binding* and two constraints which describe the use of reflexive pronouns and anaphors (personal pronouns and definite noun phrases). These constraints cover approximately the same phenomena as the binding theory of GB (Chomsky (1981); for a computational treatment, cf. Correa (1988)).

Dependency structures, by definition, refer to the sentence level of linguistic description only. The relation of dependency holds between a lexical head and one or several modifiers of that head, such that the occurrence of a head allows for the occurrence of one or several modifiers (in some pre-specified linear ordering), but *not* vice versa. Speaking in terms of dependency structure representations, the head always precedes and, thus, (transitively) *governs* its associated modifiers in the dependency tree. This basic notion of government must be further refined for the description of anaphoric relations in dependency trees (we do not claim a universal status for the following constraints, but restrict their validity to the description of the German language):

**D-binding:** A modifier $M$ is *d-bound* by some head $H$, if no node $N$ intervenes between $M$ and $H$ for which one of the following conditions holds:
(i)  node $N$ represents a finite verb, or
(ii) node $N$ represents a noun with a possessive modifier, i.e., possessive determiners, Saxon genitive, genitival and prepositional attributes.

Based on the definition of *d-binding*, we are able to specify several constraints on reflexive pronouns and anaphors in DG terms:

**Reflexive pronoun:**
The reflexive pronoun and the antecedent to which the reflexive pronoun refers are d-bound by the same head.

**Pronominal and nominal anaphors:**
(i) The antecedent $\alpha$ to which an anaphor $\pi$ refers must not be governed by the same head which d-binds $\pi$, unless (ii) applies.
(ii) The antecedent $\alpha$ to which an anaphor $\pi$ refers may *only* be governed by the same head $H1$ which d-binds $\pi$, if $\alpha$ is a modifier of a head $H2$, $H2$ is governed by $H1$, and $\alpha$ precedes $\pi$ in the linear sequence of text items. Hence, $\alpha$ is *not d-bound* by the head $H1$ which d-binds $\pi$.[1]

We will now illustrate the working of these constraints, starting with the consideration of reflexives. Usually, the antecedent of a reflexive pronoun is the subject of the clause to which the reflexive belongs. In (1), the subject *Maria* is d-bound by the same head as the reflexive *sich*.

(1) Maria$_i$ wäscht sich$_i$.
    [*Mary$_i$ washes herself$_i$.*]

Of course, the government relation between antecedent and reflexive need not be an immediate one. For instance, a preposition may occur between reflexive and verb, since the notion of *d-binding* does not discriminate between NPs and PPs (cf. (2)). If the finite verb is a modal or auxiliary verb, one or more non-finite verbs may occur between the reflexive and the finite verb (cf. (3)).

---

[1] The definition of *d-binding* roughly corresponds to the *governing category* in GB terminology, which relies upon the notion of *c-command*, while the latter two grammar constraints correspond to the three major binding principles of GB.

(2) Maria$_i$ lacht über sich$_i$.
   *[Mary$_i$ laughs about herself$_i$.]*

(3) Maria$_i$ möchte sich$_i$ verbessern.
   *[Mary$_i$ wants to improve herself$_i$.]*

If an intermediate node occurs between reflexive and antecedent which denotes a noun with a possessive modifier, the reflexive is d-bound by this noun (cf. (4) vs. (5)); hence, that modifier is the antecedent of the reflexive. Though *Maria* is the subject of (4), only *Peter* can be considered the antecedent of the reflexive, since it is d-bound by the head which d-binds *Peter*, viz. *Geschichte* (cf. Fig. 1). If the intermediate noun has no possessive modifiers, the subject of the entire clause is the antecedent of the reflexive, since the reflexive is d-bound by the finite verb irrespective of the occurrence of the (object) NP (cf. (6)).

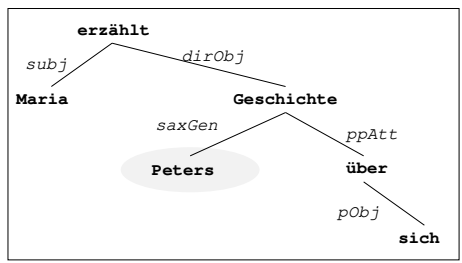

Figure 1: Dependency Tree for Examples 4 and 5

(4) Maria erzählt Peters$_i$ Geschichte über sich$_i$.
   *[Mary tells Peter's$_i$ story about himself$_i$.]*

(5) * Maria$_i$ erzählt Peters Geschichte über sich$_i$.
   *[ * Mary$_i$ tells Peter's story about herself$_i$.]*

(6) Maria$_i$ erzählt eine Geschichte über sich$_i$.
   *[Mary$_i$ tells a story about herself$_i$.]*

We will now consider constraints on intra-sentential anaphora (personal pronouns and definite NPs). As a general rule, the anaphor must not occupy the position of the reflexive pronoun. Hence, for all the examples given above, any of those sentences becomes ungrammatical if the reflexives are replaced by non-reflexive anaphoric expressions (cf. (7) vs. (2)).

(7) * Maria$_i$ lacht über sie$_i$.
   *[ * Mary$_i$ laughs about her$_i$.]*

It is also obvious that whenever the anaphor belongs to a clause which is subordinate to one that contains the antecedent, both may be coreferent: this holds independently of the ordering of antecedent and pronoun (cf. (8) vs. (11)). On the other hand, if the anaphor belongs to the matrix clause and the antecedent to the subordinate clause, coreference is excluded (cf. (9)). But one can easily think of cases where this rule is overridden. Consider, e.g., a subordinate clause preceding its matrix, as is always true for topicalizations.

The claim that a pronoun in the superordinate clause must not be coreferent to an antecedent in a subordinate clause is then obviously false (cf. (10) and (11)). In (10), the antecedent *Peter* is not d-bound by the head which d-binds the anaphor *er*, and *Peter* precedes *er*. Therefore, coreference is possible.[2]

(8) Peter$_i$ erwartet, daß er$_i$ einen Brief bekommt.
   *[Peter$_i$ expects that he$_i$ will get a letter.]*

(9) * Er$_i$ erwartet, daß Peter$_i$ einen Brief bekommt.
   *[ * He$_i$ expects that Peter$_i$ will get a letter.]*

(10) Daß Peter$_i$ einen Brief bekommt, erwartet er$_i$.
   *[That Peter$_i$ will get a letter, he$_i$ expects.]*

(11) Daß er$_i$ einen Brief bekommt, erwartet Peter$_i$.
   *[That he$_i$ will get a letter, Peter$_i$ expects.]*

Another special case arises if the antecedent is a modifier of the subject of a sentence. In this case the antecedent of a pronoun may be governed by the head which d-binds the pronoun. In (12) the pronoun belongs to the subordinate clause, but in (13) the antecedent of the pronoun belongs to the subordinate clause, and the example seems to be acceptable. In (14), where the subject *Vater* is modified by the genitive attribute *des Gewinners*, the antecedent is governed by the head which also d-binds the pronoun. Both the relative clause and the genitive attribute are modifiers of the subject, which usually occurs at the first position in the German main clause. In this case, the antecedent precedes the anaphor. Hence, coreference between anaphor and antecedent must be granted.[3]

(12) Der Mann, der sie$_i$ kennt, grüßt die Frau$_i$.
   *[The man who knows her$_i$ greets the woman$_i$.]*

(13) Der Mann, der die Frau$_i$ kennt, grüßt sie$_i$.
   *[The man who knows the woman$_i$ greets her$_i$.]*

(14) Der Vater des Gewinners$_i$ gratuliert ihm$_i$.
   *[The winner's$_i$ father congratulates him$_i$.]*

The incorporation of an ordering constraint is even more justified if one looks at sentences which have a similar structure, but are different with respect to word order (cf. (15) vs. (16)). In (15), the subordinate clause immediately follows its head word, while in (16) the subordinate clause is extraposed. In (16) the anaphor precedes its antecedent, which is governed by the head that d-binds the anaphor. This violates the given constraints, hence coreference is excluded.

---
[2] GB explains topicalization with a move of the topicalized CP into the SpecComp phrase of the highest CP, so that the pronoun does not *c-command* its antecedent (in these cases movements into an $\bar{A}$-position are assumed for which the binding principles of GB do not apply).

[3] GB explains this phenomenon by linking the modifiers to the subject as adjuncts. In this position, the pronoun does not *c-command* the antecedent, and the adjunct of the subject is also in an $\bar{A}$-position.

(15) Die Frage, ob Peter$_i$ nach Dublin fahren sollte, konnte er$_i$ noch nicht beantworten.
*[The question, whether Peter$_i$ should go to Dublin, he$_i$ couldn't decide.]*

(16) * Die Frage konnte er$_i$ noch nicht beantworten, ob Peter$_i$ nach Dublin fahren sollte.
*[* The question he$_i$ couldn't decide, whether Peter$_i$ should go to Dublin.]*

Structural constraints are necessary conditions, but additional criteria have to be considered when determining the antecedent of an anaphor. Morphosyntactic conditions require that a pronoun must agree with its antecedent in gender, number and person, while a definite NP must agree with its antecedent in number only. Moreover, conceptual criteria have to be met as in the case of nominal anaphors which must subsume their antecedents at the conceptual level. Similarly, for pronominal anaphors the selected antecedent must be permitted in those conceptual roles connecting the pronominal anaphors and its grammatical head.

The DG constraints for the use of reflexives and intra-sentential anaphora cover approximately the same phenomena as GB, but the structures used by DG analysis are less complex than those of GB and do not require the formal machinery of empty categories, binding chains and complex movements (cf. Lappin and McCord (1990, p.205) for a similar argument). Hence, our proposal provides a more tractable basis for implementation.

## 3 Major Grammatical Predicates

The *ParseTalk* model of DG (Hahn et al., 1994) exploits inheritance as a major abstraction mechanism. The entire lexical system is organized as a hierarchy of lexical classes ($isa_C$ denoting the subclass relation among lexical classes), with concrete lexical items forming the leave nodes of the corresponding lexicon grammar graph. Valency constraints are attached to each lexical item, on which the local computation of concrete dependency relations between a head and its associated modifier is based. These constraints incorporate categorial knowledge about word classes and morphosyntactic knowledge involving complex feature terms as used in unification grammars.

The definition of the grammatical predicates below is based on the following conventions: $\sqcup$ denotes the unification operation, $\bot$ the inconsistent element. Let $u$ be a complex feature term and $l$ a feature, then the extraction $u \backslash l$ yields the value of $l$ in $u$. By definition, $u \backslash l$ gives $\bot$ in all other cases. In addition, we supply access to conceptual knowledge via a KL-ONE-style classification-based knowledge representation language. The concept hierarchy consists of a set of concept names $\mathcal{F} = \{\text{COMPUTERSYSTEM, NOTEBOOK, MOTHERBOARD, ...}\}$ and a subclass relation $isa_F = \{(\text{NOTEBOOK, COMPUTERSYSTEM}), (\text{PCI-MOTHERBOARD, MOTHERBOARD}), ...\} \subset \mathcal{F} \times \mathcal{F}$. $roles \subset \mathcal{F} \times \mathcal{F}$ is the set of relations with role names $\mathcal{R} = \{has\text{-}part, has\text{-}cpu, ...\}$ and denotes the *established* relations in the knowledge base, while $\mathcal{R}$ characterizes the labels of *admitted* conceptual relations. The relation $permit \subset \mathcal{F} \times \mathcal{R} \times \mathcal{F}$ characterizes the range of possible conceptual relations among concepts, e.g., (MOTHERBOARD, $has\text{-}cpu$, CPU) $\in permit$. Furthermore, *object.attribute* denotes the value of the property *attribute* at *object* and the symbol self refers to the current lexical item. The *ParseTalk* specification language, in addition, incorporates topological primitives for relations within dependency trees. The relations left and head denote "$x$ occurs left of $y$" and "$x$ is head of $y$", resp. These primitive relations can be considered declarative equivalents to the procedural specifications used in several tree-walking algorithms for anaphora resolution, e.g., by Hobbs (1978) or Ingria and Stallard (1989). Note that in the description below rel$^+$ and rel$^*$ denote the transitive and transitive/reflexive closure of a relation rel, respectively.

```
x d-binds y :⇔
(x head⁺ y)
∧ ¬∃ z: ((x head⁺ z) ∧ (z head⁺ y)
    ∧ (z isa_C* finiteVerb
      ∨ ∃u: (z head u
        ∧ ((z spec u ∧ u isa_C* DetPossessive)
          ∨ (z saxGen u ∧ u isa_C* Noun)
          ∨ (z ppAtt u ∧ u isa_C* Noun)
          ∨ (z genAtt u ∧ u isa_C* Noun)))))
```

Box 1: *D-binding*

The possible antecedents that can be reached via anaphoric relations are described by the predicates *isPotentialReflexiveAntecedentOf* (cf. Box 2) and *isPotentialAnaphoricAntecedentOf* (cf. Box 3). These incorporate the predicate *d-binds* (cf. Box 1) which formally defines the corresponding notion from Section 2. The evaluation of the ma-

```
x isPotentialReflexiveAntecedentOf y :⇔
∃ z: (z d-binds y ∧ z d-binds x)
```

Box 2: *isPotentialReflexiveAntecedentOf*

jor predicate, *isPotentialAnaphoricAntecedentOf* (cf. Box 3), determines the candidate set of possible antecedents for (pro)nominal anaphors, and

```
x isPotentialAnaphoricAntecedentOf y :⇔
¬∃ z: (z d-binds x ∧ z d-binds y)
∧ (x left⁺ y
   ∨ ¬∃ u: (u d-binds y
         ∧ (u head⁺ x)))
```

Box 3: *isPotentialAnaphoricAntecedentOf*

thus characterizes the notion of *reachability* in formal terms. The use of constraints as filters becomes evident through the further restriction of this set by the predicates adapted to particular grammatical relations, thus taking the notion of *satisfiability* into account. For instance, the predicate *PronAnaphorTest* from Box 4 contains the grammatical constraint for pronominal anaphors according to which some pronoun and its antecedent must agree in gender, number, and person, and the conceptual constraint described in Section 2. The predicate *NomAnaphorTest* from Box 5 captures the conceptual constraint for nominal anaphors such that the concept to which the antecedent refers must be subsumed by the concept to which the anaphoric noun phrase refers. Additionally it tests whether the definite NP agrees with the antecedent in number. These two predicates cover the knowledge related to the resolution of *intra-sentential* as well as *inter-sentential* anaphora. Note the equivalence of grammatical and conceptual conditions within a single constraint. All these predicates form part of the computation process aiming at the resolution of anaphora as described in Section 4.

```
PronAnaphorTest (pro, ante):⇔
ante isa_C* Noun ∧
((pro.features\self\agr\gen)
  ⊔(ante.features\self\agr\gen) ≠ ⊥) ∧
((pro.features \self\agr\num)
  ⊔(ante.features\self\agr\num) ≠ ⊥) ∧
((pro.features\self\agr\pers)
  ⊔(ante.features\self\agr\pers) ≠ ⊥) ∧
∀x ∀ role ∈ R:
  (x head pro ∧
   (x.concept, pro.concept) ∈ roles
   ⇒(x.concept, role, ante.concept) ∈ permit)
```

Box 4: *PronAnaphorTest*

```
NomAnaphorTest (defNP, ante):⇔
ante isa_C* Noun ∧
((defNP.features \self\agr\num)
  ⊔(ante.features\self\agr\num) ≠ ⊥) ∧
ante.concept isa_F* defNP.concept
```

Box 5: *NomAnaphorTest*

## 4 Resolution of Anaphora

The *ParseTalk* environment builds on the actor computation model (Agha and Hewitt, 1987) as background for the procedural interpretation of lexicalized dependency specifications in terms of so-called word actors (cf. Schacht et al. 1994; Hahn et al. 1994). Word actors combine object-oriented features with concurrency yielding strict lexical distribution and distributed computation in a methodologically clean way. The model assumes word actors to communicate via asynchronous message passing. An actor can send messages only to other actors it knows about, its so-called acquaintances. The arrival of a message at an actor is called an event; it triggers the execution of a method that is composed of atomic actions – among them the evaluation of grammatical predicates. As we will show, the specification of a particular message protocol corresponds to the treatment of fairly general linguistic tasks, such as establishing dependencies, properly arranging coordinations, and, of course, resolving anaphors. Consequently, any of these subprotocols constitutes part of the grammar specification proper.

We shall illustrate the linguistic aspects of word actor-based parsing by introducing the basic data structures for text-level anaphora as acquaintances of specific word actors, and then turn to the general message-passing protocol that accounts for intra- as well as inter-sentential anaphora. Our exposition builds on the well-known focusing mechanism (Sidner, 1983; Grosz and Sidner, 1986). Accordingly, we distinguish each sentence's unique *focus*, a complementary list of alternate *potential foci*, and a *history list* composed of discourse elements not in the list of potential foci, but occurring in previous sentences of the current discourse segment. These data structures are realized as acquaintances of sentence delimiters to restrict the search space beyond the sentence to the relevant word actors.

The protocol level of analysis encompasses the procedural interpretation of the declarative constraints given in Section 2. At that level, in the case of reflexive pronouns, the search for the antecedent is triggered by the occurrence of a reflexive pronoun in the text. Upon instantiation of the corresponding word actor, a *SearchRefAntecedent* message will be issued. The distribution strategy of the message incorporates the syntactic restrictions for the appearance of a reflexive pronoun and its possible antecedent. This can be described in terms of two phases:

1. In *phase 1* the message is forwarded from its initiator to the word actor which *d-binds* the

initiator. Only if the message reaches a finite verb or a noun which has a possessive modifier is a new message with *phase 2* sent, and the message in *phase 1* terminates. On any other occasion (e.g., the head of the initiator is a preposition or a non-finite verb) the message is simply passed on to the receiver's head.

2. In *phase 2* the message is forwarded to the subject of the finite verb or, if a noun *d-binds* the reflexive, to the possessive modifier of the noun.

Only nouns or personal pronouns are capable of responding to *SearchRefAntecedent* messages. If the search for an antecedent is successful, a *RefAntecedentFound* message is sent directly to the initiator of the search message which changes its concept identifier accordingly.

For pronominal anaphors, the search for the antecedent is triggered by the occurrence of a personal pronoun. Upon instantiation of the corresponding word actor, a *SearchPronAntecedent* message will be sent. For nominal anaphors, the search for the antecedent is triggered by the attachment of a definite article as a modifier to its head noun, so that a *SearchNomAntecedent* message will be issued. Since the structural criteria for the sentence position of both types of anaphors are the same, the distribution mechanisms underlying the corresponding messages can be described by their common superclass, *SearchAntecedent*. Its distribution strategy incorporates the syntactic restrictions for the appearance of both elements involved, anaphor and antecedent. This can be described in terms of three main phases:

1. In *phase 1*, the message is forwarded from its initiator to the head which d-binds the initiator. Only if the message reaches this head are two further messages with *phases 1a* and *2* sent simultaneously, and the message in *phase 1* terminates. On any other occasion (e.g., the head of the pronoun is a preposition) the message is simply passed on to the receiver's head.
   (a) In *phase 1a* the modifiers of the initiator's direct head are tested, in order to determine if any of these modifiers have modifiers themselves. When the test succeeds, the message is forwarded to these modifiers, where the anaphor predicates (*PronAnaphorTest* or *NomAnaphorTest*) are evaluated in parallel.

2. In *phase 2* the message is forwarded from the head which d-binds the initiator (the original sender) to the word actor which represents the sentence delimiter of the *current* sentence. If on that path the message encounters a head which d-binds the sender (mediating messages from the initiator), that head may possibly govern an antecedent in its subtree. New messages with *phase 2a* are sent (their number depends on how many modifiers of the head exist).
   (a) In *phase 2a* the message is forwarded from the head which d-binds the sender to each of its modifiers (excluding the sender of the message), where both anaphor predicates are evaluated.

3. *Phase 3* is triggered independently from *phase 1* and *2*. The path leads from the initiator to the sentence delimiter of the *previous* sentence, where its state is set to *phase 3a*.
   (a) In *phase 3a* the sentence delimiter's acquaintances *Focus* and *PotFoci* are tested for the anaphor predicates.

Note that only nouns or personal pronouns are capable of responding to *SearchAntecedent* messages and test whether they fulfill the required criteria for an anaphoric relation. If any of the anaphor predicates succeeds, the determined antecedent sends an *AntecedentFound* message directly to the initiator of *SearchAntecedent*; this message carries the concept identifier of the antecedent. The initiator of the *SearchAntecedent* message, viz. the anaphor, upon receipt of the *AntecedentFound* message changes its concept identifier accordingly. This update of the concept identifier is the final result of anaphora resolution, a change which accounts for the coreference between concepts denoted by different lexical items at the text level.

We now discuss the protocol for establishing anaphoric relations based on intra- and intersentential anaphora considering the following text:

(17) Die Firma Compaq, die den LTE-Lite entwikkelt, bestückt *ihn* mit einem PCI-Motherboard. *Der Rechner* hat eine Taktfrequenz von 50 Mhz. *[The company Compaq, which develops the LTE-Lite, equips it with a PCI-motherboard. The system comes with a clock frequency of 50Mhz.]*

In the first sentence of (17), the *SearchAntecedent* message is caused by the occurrence of the personal *ihn* (cf. Fig. 2 which depicts two instances of anaphora resolution). In *phase 1*, the message reaches the finite verb *bestückt*, where two new instances of the message are created. In *phase 2* it takes the path to the sentence delimiter of the current sentence (no effect). In *phase 1a*, the message reaches the subject *Firma*, which is the leftmost modifier of the verb, and determines the noun *LTE-Lite* as the only possible antecedent of *ihn*. The success of *PronAnaphorTest* leads to the sending of an *AntecedentFound* message, the result of which is the update of the concept identi-fier of *ihn* with that of *LTE-Lite*. Simultaneously, a *SearchPronAntecedent* message in *phase 3* takes the path to the sentence delimiter of the previous sentence, where it evaluates *PronAnaphorTest* with respect to its acquaintances *Focus* and *PotFoci* (no effect).

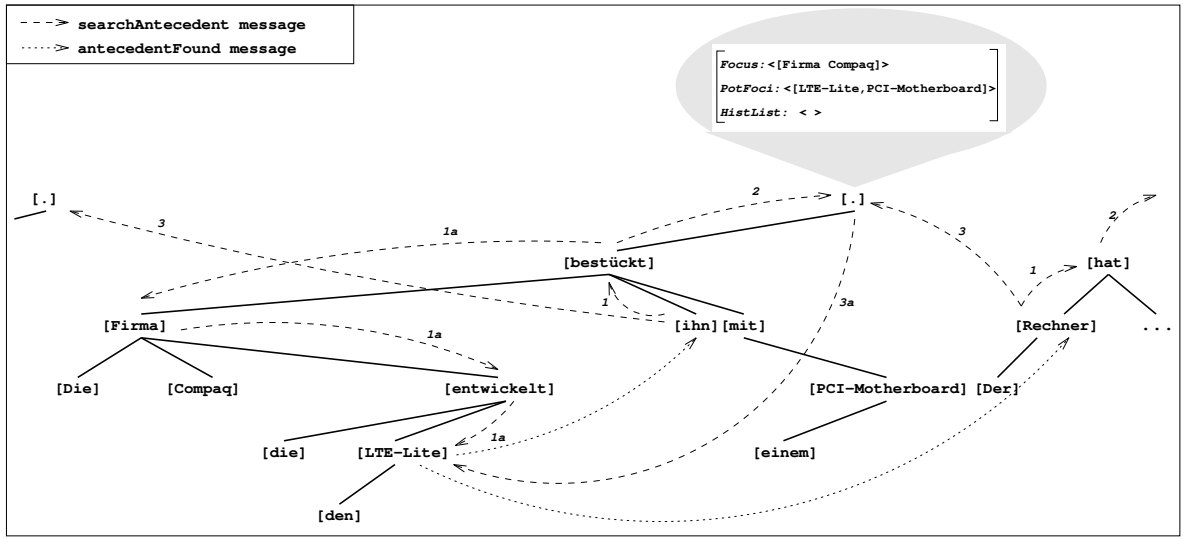

Figure 2: Sample Communication Protocol

The second sentence of (17) contains the definite noun phrase *der Rechner*. The search of an antecedent is triggered by the attachment of the definite article to the noun. In *phase 1* the message reaches the finite verb *hat*, where new instances of the message are created. *Phase 1a* yields no positive results and the message terminates. In *phase 2* the message takes the path from the finite verb to the sentence delimiter (no effect). Since there are no possible antecedents within the sentence, in *phase 3* possible antecedents are checked which are stored as the acquaintances *Focus* and *PotFoci* of the sentence delimiter for the previous sentence. Since *Rechner* subsumes *LTE-Lite* at the conceptual level, *NomAnaphorTest* succeeds. An *AntecedentFound* message is created, which changes the concept identifier of *Rechner* appropriately.

## 5 Comparison to Related Work

From the linguistic viewpoint, sentence anaphora, so far, have only been sketchily dealt with by dependency grammarians, e.g., by Hudson (1984; 1990). The most detailed description of grammatical regularities and an associated parsing procedure has been supplied by Lappin and McCord (1990). It is based on the format of a slot grammar (SG), a slight theory variant of DG. In particular, they treat pronominal coreference and anaphora (i.e., reflexives and reciprocals). Our approach methodologically differs in three major aspects from that study: First, unlike the SG proposal, which is based on a second-pass algorithm operating on fully parsed clauses to determine anaphoric relationships, our proposal is basically an incremental single-pass parsing model. Most important, however, is that our model incorporates the text-level of anaphora resolution, a shortcoming of the original SG approach that has recently been removed (Lappin and Leass, 1994), but still is a source of lots of problems. Third, unlike our approach, even the current SG model for anaphora resolution does not incorporate conceptual knowledge and global discourse structures (for reasons discussed by Lappin and Laess). This decision might nevertheless cause trouble if more conceptually rooted text cohesion and coherence structures have to be accounted for (e.g., textual ellipses).

A particular problem we have not yet solved, the plausible ranking of single antecedents from a candidate set, is dealt with in depth by Lappin and Laess (1994) and Hajicova et al. (1992). Both define salience metrics capable of ordering alternative antecedents according to structural criteria, several of which can directly be attributed to the topological structure and topic/comment annotations of the underlying dependency trees.

# 6 Conclusions

We have outlined a model of anaphora resolution which is founded on a dependency-based grammar model. This model accounts for sentence-level anaphora, with constraints adapted from GB, as well as text-level anaphora, with concepts close to Grosz-Sidner-style focus models. The associated text parser is based on the actor computation model. Its message passing mechanisms constitute the foundation for expressing specific linguistic protocols, e.g., that for anaphora resolution. The main advantage of our approach lies in the unified framework for sentence- and text-level anaphora, using a coherent grammar format, and the provision for access to grammatical and conceptual knowledge without prioritizing either one of them. It is also a striking fact that, given the same linguistic phenomena, structural dependency configurations are considerably simpler than their GB counterparts, though suitably expressive.

The anaphora resolution module (for reflexives, intra- and inter-sentential anaphora) has been realized as part of *ParseTalk*, a dependency parser which forms part of a larger text understanding system for the German language, currently under development at our laboratory. The parser has been implemented in Smalltalk; the Smalltalk system itself, which runs on a SUN SparcStation network, has been extended by asynchronous message passing facilities and physical distribution mechanisms (Xu, 1993). The current lexicon contains a hierarchy of approximately 100 word class specifications with nearly 3.000 lexical entries and corresponding concept descriptions from two domains (information technology and medicine) available from the LOOM knowledge representation system (MacGregor and Bates, 1987).

**Acknowledgments.** We would like to thank our colleagues in the $\mathcal{CLIF}$ group who read earlier versions of this paper. In particular, improvements are due to discussions we had with S. Schacht, N. Bröker, P. Neuhaus, and M. Klenner. We also like to thank J. Alcantara (Cornell U) who kindly took the role of the native speaker via Internet. This work has been funded by *LGFG Baden-Württemberg* (1.1.4-7631.0; M. Strube) and a grant from *DFG* (Ha 2907/1-3; U. Hahn).